\documentclass[jou,floatsintext]{apa7}

\usepackage[american]{babel}
\usepackage[utf8]{inputenc}
\usepackage{float}
\usepackage[T1]{fontenc}
\usepackage{csquotes} 
\usepackage[style=apa,backend=biber,natbib=true]{biblatex}
\DeclareLanguageMapping{american}{american-apa} 
\usepackage{caption}
\usepackage{graphicx}

\usepackage{csquotes}      
\usepackage{amsmath}       
\usepackage{multirow}

\title{Automated Alignment of Math Items to Content Standards in Large-Scale Assessments Using Language Models}
\shorttitle{Automated Alignment Using Language Models}
\leftheader{Automated Alignment Using Language Models}

\author{Qingshu Xu\textsuperscript{1,2}, Hong Jiao\textsuperscript{1}, Tianyi Zhou\textsuperscript{1}, Ming Li\textsuperscript{1}, Nan Zhang\textsuperscript{1},Sydney Peters\textsuperscript{1}, Yanbin Fu\textsuperscript{1}}

\affiliation{
    \textsuperscript{1}University of Maryland, College Park\\
    \textsuperscript{2}Shandong Jiaotong University
}

\abstract{%
Accurate alignment of items to content standards is critical for valid score interpretation in large-scale assessments. This study evaluates three automated paradigms for aligning items with four domain and nineteen skill labels. First, we extracted embeddings and trained multiple classical supervised machine learning models, and further investigated the impact of dimensionality reduction on model performance. Second, we fine-tuned eight BERT model and its variants for both domain and skill alignment. Third, we explored ensemble learning with majority voting and stacking with multiple meta-models. The DeBERTa-v3-base achieved the highest weighted-average F1 score of 0.950 for domain alignment while the RoBERTa-large yielded the highest F1 score of 0.869 for skill alignment. Ensemble models did not surpass the best-performing language models. Dimension reduction enhanced linear classifiers based on embeddings but did not perform better than language models. This study demonstrated different methods in automated item alignment to content standards.}

\begin{document}
\maketitle

\section{Introduction}

In educational assessments, alignment of  items to content standards is fundamental to establishing score validity.
Traditionally, item alignment relies on subject–matter experts (SMEs) manually reviewing and matching items to content standards \citep{Webb1999}, a labor-intensive task prone to fatigue and semantic ambiguity \citep{ButterfussDoran2025,Camilli2024}. 
Recent advances in natural language processing (NLP) have dramatically improved text understanding and processing for prediction or classification. 
In recent years, embedding-based auto-alignment methods have leveraged distributed word representations to map assessment items to content standards. 
Early work demonstrated that static embeddings from Word2Vec \citep{Mikolov2013}, GloVe \citep{Pennington2014}, and FastText \citep{Joulin2017}, can capture latent semantic features by representing each token as a dense vector in a high-dimensional space. 
These vectors are typically averaged or pooled over an item’s text to produce a single feature representation, enabling similarity measures or clustering for automatic alignment. 
For example, Zhou and Ostrow showed that averaging word-level embeddings and applying cosine similarity achieved up to 35\% accuracy when predicting alignments between standards and item texts, with contextual sentence embeddings yielding further gains \citep{ZhouOstrow2022}. 
However, static embeddings are context‐ignorant, assigning each word a single, fixed vector regardless of its surrounding context, so they cannot capture polysemy or nuanced syntactic relationships, which often leads to semantic mismatches in fine‐grained alignment tasks.

The advent of contextualized embeddings and transformer-based models has markedly improved automated item alignment performance. 
Models such as ELMo \citep{Ethayarajh2019}, BERT \citep{Devlin2019}, and XLNet incorporate contextual dependencies, allowing polysemous tokens to be represented differently by position. 
Devlin et al. introduced a deep bidirectional transformer that, when fine-tuned on limited labeled data, outperforms static embeddings across diverse text-classification tasks \citep{Devlin2019}. 
Reimers and Gurevych further proposed Sentence-BERT, which uses Siamese BERT networks to generate semantically rich sentence embeddings, improving clustering and retrieval applications in educational contexts \citep{Reimers2019}.

Building on these advances, several studies have applied transformer models directly to alignment tasks. Catalog \citep{Khan2021Catalog} uses a transformer-based semantic-matching algorithm to rank educational materials against Next Generation Science Standards (NGSS), reducing expert review by prioritizing top candidates. 
The Semantic-Knowledge Mapping Network (S-KMN) integrates both static and contextual features within a multi-channel architecture to annotate quiz questions with curriculum labels, achieving good results on large quiz datasets \citep{Wang2023}. 
Collectively, these embedding- and transformer-based approaches highlight the potential for automated alignment tools to streamline item alignment workflows and support, rather than replace expert judgment.

To date, little research has undertaken a systematic comparison of embedding-based, transformer-based, and the ensemble of different approaches to automated item content alignment of large-scale test items. A comprehensive review thus motivates the present study on automated alignment of large-scale test items to content standards. To guide a systematic investigation of automated item alignment to content standards, this study addresses the following research questions:

\begin{enumerate}
  \item \textbf{Embedding-based models:} How accurately can embedding-based supervised machine learning models, when used with cosine similarity, align items to content standards?
  \item \textbf{Dimensionality reduction:} Can dimensionality reduction using Principal Component Analysis (PCA) on embeddings improve the performance of supervised machine learning models?
  \item \textbf{Fine‐Tuned Transformer-based models:} To what extent do BERT and its variant models differ in performance when finetuned for item alignment?
  \item \textbf{Ensemble Learning:} Do ensemble strategies of majority voting and stacking with embedding- and Transformer-based base learners yield better performance than the best fine-tuned language models?
\end{enumerate}

\section{Methods}
\subsection{Datasets}

The dataset comprises 1,385 Math items from the question bank of a large-scale mathematics test. 
In our data‐preparation pipeline, each Math item is encoded as a self‐contained JSON object whose fields mirror the visual and semantic structure of the original question. 
Metadata such as question ID, assessment name, test subject, domain, skill, and difficulty are captured in top‐level keys (e.g. “Question ID”, “Domain”, “Skill”, “Question Difficulty”), enabling straightforward filtering by content category. 
The main prompt is stored under “Question Text” with all mathematical notation preserved in LaTeX (e.g. “\$(d,4)\$”) to ensure unambiguous formula reconstruction. 
For multiple-choice items, each answer choice (“Choice A”–“Choice D”) retains any fraction or symbol in escaped LaTeX form. 
The correct answer key simply records the letter (e.g. `"C"`), while the full problem solution appears in a “Rationale” field, again leveraging LaTeX for all algebraic steps. 
Non‐textual elements are handled by two optional fields: “Table” holds a latex rendering of any structured data tables, and “Figure” contains an accessible, prose‐based description of graphs or diagrams (e.g. “line passes through (0,7) and (8,0)”). 

This schema affords modularity—downstream processes can concatenate all text fields to build token sequences—provides clarity by mapping visual layout to JSON keys, and ensures full preservation of mathematical semantics for both human review and machine processing.
Each item is annotated with one of four domains (Advanced Math: 394 items; Algebra: 457 items; Geometry and Trigonometry: 195 items; Problem-Solving and Data Analysis: 339 items) and one of nineteen skill labels (e.g., Linear functions; Systems of two linear equations in two variables). 

When these fields are concatenated, token counts (whitespace-delimited) range from 36 to 811 tokens, with an average of approximately 213.6 and a median of 197. 
Domain-level average token counts span from 200.2 (Advanced Math) to 241.3 (Geometry and Trigonometry), as shown in Table \ref{tab:domain_stats}.
\begin{table}[ht]
  \caption{Descriptive Statistics of Tokens by Domains}
  \label{tab:domain_stats}
  \centering
  \resizebox{\columnwidth}{!}{%
    \begin{tabular}{lccc}
      \hline
      \textbf{Domain} & \textbf{Count} & \textbf{Avg.\ Tokens} & \textbf{Median Tokens} \\
      \hline
      Advanced Math                     & 394 & 200.22 & 172 \\
      Algebra                           & 457 & 209.18 & 200 \\
      Geometry and Trigonometry         & 195 & 241.28 & 222 \\
      Problem-Solving and Data Analysis & 339 & 219.06 & 207 \\
      \hline
    \end{tabular}%
  }
\end{table}

Skill‐level item counts range from 10 items in “Evaluating statistical claims: Observational studies and experiments” to 179 items in “Nonlinear functions,” while mean combined‐token lengths vary from approximately 90.5 tokens for “Equivalent expressions” to 321.5 tokens for “Two-variable data: Models and scatterplots,” reflecting considerable heterogeneity in textual complexity across skills (see Table \ref{tab:skill_stats}).
\begin{table}[ht]
  \caption{Descriptive Statistics of Tokens by Skills.}
  \label{tab:skill_stats}
  \centering
  \resizebox{\columnwidth}{!}{%
    \begin{tabular}{lrrr}
      \hline
      \textbf{Skill} & \textbf{Count} & \textbf{Avg.\ Tokens} & \textbf{Median Tokens} \\
      \hline
      Area and volume                                                     &  60 & 207.15 & 193.50 \\
      Circles                                                             &  35 & 251.40 & 243.00 \\
      Equivalent expressions                                              &  92 & 141.45 & 126.00 \\
      Evaluating statistical claims: Observational studies and experiments &  10 & 278.60 & 269.00 \\
      Inference from sample statistics and margin of error                &  22 & 261.14 & 264.50 \\
      Linear equations in one variable                                    &  83 & 143.04 & 126.00 \\
      Linear equations in two variables                                   & 101 & 228.64 & 219.00 \\
      Linear functions                                                    & 125 & 201.66 & 193.00 \\
      Linear inequalities in one or two variables                         &  58 & 226.02 & 212.00 \\
      Lines, angles, and triangles                                        &  62 & 193.35 & 182.00 \\
      Nonlinear equations in one variable and systems of equations in two variables & 123 & 185.81 & 164.00 \\
      Nonlinear functions                                                 & 179 & 240.34 & 218.00 \\
      One-variable data: Distributions and measures of center and spread   &  66 & 244.94 & 227.00 \\
      Percentages                                                         &  67 & 155.90 & 142.00 \\
      Probability and conditional probability                             &  41 & 216.10 & 207.00 \\
      Ratios, rates, proportional relationships, and units                &  75 & 154.84 & 138.00 \\
      Right triangles and trigonometry                                    &  38 & 256.39 & 225.00 \\
      Systems of two linear equations in two variables                    &  90 & 226.11 & 207.50 \\
      Two-variable data: Models and scatterplots                          &  58 & 321.50 & 310.50 \\
      \hline
    \end{tabular}%
  }
\end{table}

This dataset offers several merits: rich multi-field content including stems, choices, rationales, and textual descriptions of tables, equations, and figures (rendered in words or LaTeX), provides diverse contextual cues; adequate sample size (1,385 items) and topical diversity provides a context of item alignment to a real large-scale testing program; token-length variability tests model capacity for succinct vs. verbose inputs; inclusion of non-textual elements encoded as text encourages exploration of multi-modal item alignment and domain-specific architectures.

We partitioned the full dataset of 1,385 math items into training, validation, and test sets in a 70 \% / 15 \% / 15 \% ratio using stratified sampling to preserve the relative frequencies of both domain and skill labels.
The 70 \% training set provides ample examples for model training; the 15 \% validation set is reserved for hyperparameter tuning and identification of the best-performing model; and the 15 \% test set enables an unbiased assessment of generalization performance of the best-performing model on unseen datasets.
This hold-out scheme follows common practice in supervised learning to balance training efficacy with model on unseen dataset \citep{Goodfellow2016}.

\subsection{Models}

\subsubsection{Embedding‐based Supervised Machine Learning Models}

Prior work on automated item alignment has predominantly relied on static word embeddings from Word2Vec \citep{Mikolov2013}, GloVe \citep{Pennington2014}, and FastText \citep{Joulin2017}, or contextual embeddings obtained from BERT \citep{Devlin2019} and Sentence‐BERT \citep{Reimers2019} to represent question texts.
Static embeddings assign each token a single vector irrespective of context, while vanilla BERT embeddings, though contextual, are not explicitly optimized for retrieval or cross‐item similarity.
To overcome these limitations, we employed the instruction‐tuned embedding model \texttt{intfloat/multilingual-e5-large-instruct} within the SentenceTransformers framework.  This model benefits from (1) multilingual pretraining, ensuring robust semantic alignment across diverse content, and (2) instruction tuning, which tailors the embedding space for downstream tasks such as semantic retrieval or classification.
Empirical benchmarks have demonstrated that E5‐based embeddings outperform both static and BERT‐derived embeddings on retrieval and similarity tasks \citep{wang2024multilingual}.  
These embeddings then serve as input features to train a suite of traditional machine learning classifiers including logistic regression, Naive Bayes, gradient boosting machines, support vector machines as well as other methods such as random forests and multilayer perceptrons, enabling a comprehensive comparison of performance using high‐quality, instruction‐tuned embedding representations.

Given the high dimensionality of the resulting embeddings and inspired by Gui's study \citep{gui2025efficient}, we also applied PCA to reduce the 1,024 dimensions. 
We selected components either by retaining those with eigenvalues greater than one or by using cumulative explained‐variance thresholds, and then trained the same suite of classifiers on these lower‐dimensional representations. 
This allowed us to compare performance when using raw embeddings versus PCA‐reduced components.

\subsubsection{Fine-Tuned Bert and its Variant Models}

We fine-tuned pre-trained transformer models from the Hugging Face Transformers library \citep{Wolf2020} to perform item alignment to both 4 domains and 19 skills.  
Specifically, we loaded the pre-trained BERT variants (like bert-base-uncased, RoBERTa, DeBERTa) and then fine-tuned all Transformer parameters (not just the classification head) by backpropagating the cross-entropy loss through the entire network. We attached a classification head, a single fully connected layer, on top of the pooled [CLS] representation and updated both its weights and those of the underlying BERT layers during training.

Each Math item’s combined text (stem, answer choices, rationale, and LaTeX-rendered descriptions of equations, tables and figures) was tokenized with the model’s tokenizer and formatted to a maximum sequence length of 512 tokens (padding shorter sequences, truncating longer ones). 
Models were trained separately for domain and skill labels using cross-entropy loss and the AdamW optimizer \citep{Loshchilov2019} with a learning rate of $2 \times 10^{-5}$, weight decay of 0.01,  and up to five epochs. 
We applied early stopping with a patience of five epochs on the validation set, selecting the best checkpoint based on F1 scores. By adapting contextual embeddings directly to the math item alignment task, this approach leverages self-attention to capture both local and long-range dependencies in the concatenated item text, yielding substantial performance gains over static and off-the-shelf embedding classifiers.

\subsubsection{Ensemble Learning}

Ensemble learning combines multiple base learners to improve predictive accuracy and robustness by aggregating their outputs \citep{Dietterich2000,Zhou2012}. 
Common ensemble strategies include majority voting where each model casts one “vote” and the most frequent label is chosen and stacking (stacked generalization), which trains a meta model based on the predictions of base models to correct individual biases \citep{Wolpert1992,Pedregosa2011}.  
Ensembles tend to reduce variance, mitigate overfitting, and capture complementary strengths of diverse architectures.

In this study, we first applied majority voting to the skill predictions of four fine‐tuned BERT‐family variants (BERT, ALBERT, RoBERTa, MathBERT), selecting the label with the highest vote count as the final decision.  
Next, we implemented stacking: the base‐level models comprised the best embedding‐based classifier (MLP on E5 embeddings) and the best of the BERT‐family fine‐tuned model; their probabilistic outputs from the base models were used as input features to train the meta-model. 
The stacked model was then evaluated on the held‐out test set. 
By comparing majority voting and stacking, we analyze how ensemble fusion strategies leverage heterogeneous signals from embedding‐ and transformer‐based approaches to further enhance automated item alignment performance.

\subsection{Evaluation Criteria}

We evaluate each model's performance using Precision, Recall, and F1 scores, defined in equations 1, 2, and 3 respectively. To report an overall accuracy score across all classes, we compute the weighted-average F1 scores as presented in equation 4.
\begin{equation}
\mathrm{Precision} = \frac{TP}{TP + FP}
\end{equation}
\begin{equation}
\mathrm{Recall} = \frac{TP}{TP + FN}
\end{equation}
\begin{equation}
F1 = 2 \cdot \frac{\mathrm{Precision}\times\mathrm{Recall}}{\mathrm{Precision} + \mathrm{Recall}}
\end{equation}
\begin{equation}
F1{\mathrm{weighted}}
= \sum_{c \in C} \frac{n_{c}}{\sum_{j \in C} n_{j}} \; F_{1,c}
\end{equation}
where $C$ is the set of all classes and $n_{c}$ is the number of true instances of class $c$.  

\section{Results}
\subsection{Embedding-based Models}

We first encoded each item using the instruction‐tuned E5 SentenceTransformer (\texttt{intfloat/multilingual-e5-large-instruct}). 
In the embedding‐based experiments, we concentrated on skill‐level classification (19 labels) rather than the coarser domain attribute (4 labels).  With only four broad domains, even simple centroid or rule‐based approaches yield high accuracy, offering little insight into the embedding’s representational power.  
In contrast, the 19 skill categories pose a finer‐grained, more challenging task that better exercises the discriminative capacity of the 1,024-dimensional E5 embeddings across the studied classifiers.
For each skill category, we mean‐pooled the embeddings of all items labeled with that skill to obtain prototype vectors.  
We then computed the pairwise cosine similarity among the 19 skills (see Table \ref{tab:cosine_matrix} ). 
The off‐diagonal similarities remain very high with a minimum of 0.881, a median of 0.950, a mean of 0.949, and a maximum of 0.994, indicating that items of different skills are semantically similar in the embedding space.
Some prior studies have used cosine similarity directly for alignment of standards to standards \citep{ButterfussDoran2025,ZhouOstrow2022}. 
Thus, relying solely on text similarity metrics may not be sufficient for distinguishing skill categories when aligning items to content standards.

\begin{table*}[t]
  \caption{Cosine Similarity Matrix among Skill Embeddings.}
  \label{tab:cosine_matrix}
  \centering
  \resizebox{\textwidth}{!}{%
    \begin{tabular}{lrrrrrrrrrrrrrrrrrrr}
      \hline
      \textbf{Skill} 
      & \textbf{ARV} & \textbf{CIR} & \textbf{EQX} & \textbf{ESC} & \textbf{ISM}
      & \textbf{LE1} & \textbf{LE2} & \textbf{LFN} & \textbf{LIT} & \textbf{LAT}
      & \textbf{NEQ} & \textbf{NFU} & \textbf{OVD} & \textbf{PER} & \textbf{PCP}
      & \textbf{RAT} & \textbf{RTT} & \textbf{STE} & \textbf{TVD} \\
      \hline
      ARV  & 1.00 & 0.96 & 0.94 & 0.90 & 0.93 & 0.95 & 0.95 & 0.95 & 0.94 & 1.00 
           & 0.93 & 0.95 & 0.96 & 0.95 & 0.93 & 0.96 & 0.93 & 0.92 & 0.94 \\
      CIR  & 0.96 & 1.00 & 0.94 & 0.89 & 0.92 & 0.95 & 0.96 & 0.95 & 0.94 & 0.95 
           & 0.94 & 0.95 & 0.96 & 0.94 & 0.93 & 0.96 & 0.97 & 0.93 & 0.95 \\
      EQX  & 0.94 & 0.94 & 1.00 & 0.90 & 0.93 & 0.98 & 0.96 & 0.96 & 0.94 & 0.97 
           & 0.96 & 0.97 & 0.96 & 0.96 & 0.95 & 0.97 & 0.95 & 0.93 & 0.96 \\
      ESC  & 0.90 & 0.89 & 0.90 & 1.00 & 0.92 & 0.89 & 0.90 & 0.92 & 0.89 & 0.92 
           & 0.90 & 0.92 & 0.96 & 0.93 & 0.95 & 0.94 & 0.89 & 0.93 & 0.90 \\
      ISM  & 0.93 & 0.92 & 0.93 & 0.92 & 1.00 & 0.93 & 0.93 & 0.95 & 0.93 & 0.95 
           & 0.93 & 0.95 & 0.96 & 0.97 & 0.96 & 0.96 & 0.93 & 0.95 & 0.94 \\
      LE1  & 0.95 & 0.95 & 0.98 & 0.89 & 0.93 & 1.00 & 0.98 & 0.97 & 0.94 & 0.97 
           & 0.97 & 0.98 & 0.93 & 0.97 & 0.95 & 0.97 & 0.97 & 0.95 & 0.97 \\
      LE2  & 0.95 & 0.96 & 0.96 & 0.90 & 0.93 & 0.98 & 1.00 & 0.97 & 0.97 & 0.98 
           & 0.97 & 0.98 & 0.94 & 0.97 & 0.97 & 0.96 & 0.95 & 0.97 & 0.96 \\
      LFN  & 0.95 & 0.95 & 0.96 & 0.92 & 0.95 & 0.97 & 0.97 & 1.00 & 0.97 & 0.99 
           & 0.97 & 1.00 & 0.96 & 0.94 & 0.96 & 0.97 & 0.97 & 0.97 & 0.97 \\
      LIT  & 0.94 & 0.94 & 0.94 & 0.89 & 0.93 & 0.94 & 0.97 & 0.97 & 1.00 & 0.97 
           & 0.97 & 0.97 & 0.94 & 0.97 & 0.96 & 0.97 & 0.94 & 0.96 & 0.95 \\
      LAT  & 1.00 & 0.95 & 0.97 & 0.92 & 0.95 & 0.97 & 0.98 & 0.99 & 0.97 & 1.00 
           & 0.98 & 1.00 & 0.96 & 0.97 & 0.96 & 0.94 & 0.97 & 0.95 & 0.97 \\
      NEQ  & 0.93 & 0.94 & 0.96 & 0.90 & 0.93 & 0.97 & 0.97 & 0.97 & 0.97 & 0.98 
           & 1.00 & 0.99 & 0.96 & 0.94 & 0.96 & 0.94 & 0.95 & 0.92 & 0.93 \\
      NFU  & 0.95 & 0.95 & 0.97 & 0.92 & 0.95 & 0.98 & 0.98 & 1.00 & 0.97 & 1.00 
           & 0.99 & 1.00 & 0.96 & 0.97 & 0.97 & 0.94 & 0.97 & 0.94 & 0.97 \\
      OVD  & 0.96 & 0.96 & 0.96 & 0.96 & 0.96 & 0.93 & 0.94 & 0.96 & 0.94 & 0.96 
           & 0.96 & 0.96 & 1.00 & 0.98 & 0.96 & 0.96 & 0.95 & 0.94 & 0.96 \\
      PER  & 0.95 & 0.94 & 0.96 & 0.93 & 0.97 & 0.97 & 0.97 & 0.94 & 0.97 & 0.97 
           & 0.94 & 0.97 & 0.98 & 1.00 & 0.97 & 0.95 & 0.95 & 0.94 & 0.95 \\
      PCP  & 0.93 & 0.93 & 0.95 & 0.95 & 0.96 & 0.95 & 0.97 & 0.96 & 0.96 & 0.96 
           & 0.96 & 0.97 & 0.96 & 0.97 & 1.00 & 0.95 & 0.96 & 0.94 & 0.96 \\
      RAT  & 0.96 & 0.96 & 0.96 & 0.94 & 0.96 & 0.97 & 0.96 & 0.97 & 0.97 & 0.94 
           & 0.94 & 0.94 & 0.96 & 0.95 & 0.95 & 1.00 & 0.93 & 0.95 & 0.94 \\
      RTT  & 0.93 & 0.97 & 0.95 & 0.89 & 0.93 & 0.97 & 0.95 & 0.97 & 0.94 & 0.97 
           & 0.95 & 0.94 & 0.95 & 0.95 & 0.96 & 0.93 & 1.00 & 0.92 & 0.94 \\
      STE  & 0.92 & 0.93 & 0.93 & 0.93 & 0.95 & 0.95 & 0.97 & 0.97 & 0.96 & 0.95 
           & 0.92 & 0.94 & 0.94 & 0.94 & 0.94 & 0.95 & 0.92 & 1.00 & 0.95 \\
      TVD  & 0.94 & 0.95 & 0.96 & 0.90 & 0.94 & 0.97 & 0.96 & 0.97 & 0.95 & 0.97 
           & 0.93 & 0.97 & 0.96 & 0.95 & 0.96 & 0.94 & 0.94 & 0.95 & 1.00 \\
      \hline
    \end{tabular}%
  }
  \begin{flushleft}\footnotesize
  \textbf{Notes:} ARV = Area and volume; CIR = Circles; EQX = Equivalent expressions; ESC = Evaluating statistical claims; ISM = Inference from sample statistics; LE1 = Linear equations in one variable; LE2 = Linear equations in two variables; LFN = Linear functions; LIT = Linear inequalities; LAT = Lines, angles, and triangles; NEQ = Nonlinear equations \& systems; NFU = Nonlinear functions; OVD = One‐variable data; PER = Percentages; PCP = Probability \& conditional probability; RAT = Ratios, rates, and units; RTT = Right triangles \& trigonometry; STE = Systems of two linear equations; TVD = Two‐variable data.
  \end{flushleft}
\end{table*}

We trained logistic regression, random forest (RF), gradient boosting (GB), XGBoost (XGB), LightGBM (LGBM), support vector machine (SVM), multilayer perceptron (MLP), k-nearest neighbors (KNN), and Naive Bayes (NB) classifiers on the 1,024 dimension embeddings produced by the E5 model to predict the 19 skill categories. As all precisions and recalls are larger than 0.5, only F1 scores are presented in this report.
Table~\ref{tab:embed_pca_results} reports their weighted-average F1 scores. 
The MLP achieves the highest score (0.847), demonstrating its ability to model complex, non-linear relationships in the embedding space, while logistic regression attains the lowest score (0.608) due to its linear decision boundary. 
Among tree-based ensembles, LightGBM (0.804) and XGBoost (0.786) outperform random forest (0.753) and gradient boosting (0.654) by more effectively capturing feature interactions. SVM (0.800) also performs robustly, whereas KNN (0.786) and NB (0.793) deliver moderate results, reflecting their sensitivity to local data structure and independence assumptions, respectively.

To mitigate the high dimensionality (1,024) of the E5 embeddings, we applied PCA, which can reduce noise, decorrelate features, and improve model training efficiency and generalization \citep{Jolliffe2002,Abdi2010}.  We selected principal components by four criteria: eigenvalues > 1 (Kaiser’s rule), and cumulative explained variance thresholds of 90\%, 95\%, and 99\%. Table \ref{tab:embed_pca_results} reports the weighted‐average F1 score for each reduction strategy. The best F1 score in each column is highlighted in bold.

\begin{table}[ht]
  \caption{Performance Comparison of Embedding‐based Models with and without PCA Reduction under Different Selection Criteria. }
  \label{tab:embed_pca_results}
  \centering
  \resizebox{\columnwidth}{!}{%
    \begin{tabular}{lrrrrr}
      \hline
      \textbf{Model} & \textbf{F1$_{\mathrm{emb}}$} & \textbf{F1$_{\mathrm{PCA}}$} & \textbf{F1$_{\mathrm{PCA,90\%}}$} & \textbf{F1$_{\mathrm{PCA,95\%}}$} & \textbf{F1$_{\mathrm{PCA,99\%}}$} \\
      \hline
      logistic & 0.6076 & 0.8110 & 0.8150 & \textbf{0.8354} & 0.8305 \\
      rf       & 0.7525 & 0.7736 & 0.7698 & 0.7245 & 0.6809 \\
      gb       & 0.6538 & 0.6840 & 0.6384 & 0.6290 & 0.6306 \\
      xgb      & 0.7858 & 0.7321 & 0.7211 & 0.7081 & 0.7070 \\
      lgbm     & 0.8036 & 0.7595 & 0.7417 & 0.7471 & 0.7239 \\
      svm      & 0.8004 & \textbf{0.8355} & \textbf{0.8263} & 0.8213 & \textbf{0.8310} \\
      mlp      & \textbf{0.8466} & 0.8108 & 0.8055 & 0.7941 & 0.8055 \\
      knn      & 0.7863 & 0.7811 & 0.7755 & 0.7779 & 0.7544 \\
      nb       & 0.7934 & 0.6973 & 0.6794 & 0.6255 & 0.5417 \\
      \hline
    \end{tabular}%
  }
\end{table}

Compared to the original embedding‐based models (best MLP: 0.8466; worst logistic: 0.6076), PCA yields large gains for linear models, that is, logistic regression improves by 33.5\% under Kaiser’s rule, but reduces performance for tree‐based ensembles and MLP, since aggressive dimensionality reduction can discard subtle feature interactions that these more complex learners exploit.
In particular, PCA also improved the performance of SVM: its weighted average F1 score increased from 0.8004 (without PCA) to 0.8355 under the eigenvalue->1 criterion, a relative gain of 4.39\%. This suggests that SVM’s margin-maximization benefits from the decorrelated, lower dimensional feature space produced by PCA, which filters noise and highlights the most discriminative components.

\subsection{Fine-Tuned Bert and its Variant Models}

We fine‐tuned a suite of pre‐trained Transformer encoders from the BERT family, namely \texttt{bert-base-uncased} and \texttt{bert-large-uncased} (BERT), \texttt{roberta-base} and \texttt{roberta-large} (RoBERTa), \texttt{albert-base-v2} (ALBERT), \texttt{microsoft/deberta-v3-base} and \texttt{microsoft/deberta-v3-large} (DeBERTa-v3), \texttt{tbs17/MathBERT} (MathBERT), and \texttt{google/electra-base-discriminator} and \texttt{google/electra-small-discriminator} (ELECTRA).  
Each model was initialized from its Hugging Face checkpoint, augmented with a lightweight classification head on the pooled \texttt{[CLS]} token, and fine‐tuned end‐to‐end on the math domain and skill alignment tasks.

\begin{table}[ht]
  \caption{Evaluation Statistics by Model for Domain and Skill Alignment.}
  \label{tab:bert_finetune_comparison}
  \centering
  \resizebox{\columnwidth}{!}{%
    \begin{tabular}{lrrrrrrrrrr}
      \toprule
	    \multicolumn{1}{c}{\multirow{2}*{Model Name}}\
        & \multicolumn{3}{c}{Domain}
        & \multicolumn{3}{c}{Skill} \\
      \cmidrule(lr){2-4} \cmidrule(lr){5-7}
        & F1 Score & Precision & Recall
        & F1 Score & Precision & Recall \\
      \midrule
      albert\_base                  & 0.885 & 0.902 & 0.884 & 0.712 & 0.660 & 0.687 \\
      conv-bert-base                & 0.906 & 0.911 & 0.910 & 0.692 & 0.657 & 0.654 \\
      roberta-base      & 0.899 & 0.911 & 0.906 & 0.833 & 0.770 & 0.797 \\
      roberta-large     & 0.943 & 0.939 & 0.946 & \textbf{0.869} & 0.866 & 0.845 \\
      electra-base-discr.   & 0.913 & 0.918 & 0.920 & 0.705 & 0.679 & 0.637 \\
      electra-small-discr.  & 0.869 & 0.879 & 0.856 & 0.278 & 0.232 & 0.276 \\
      bert-base-uncased      & 0.920 & 0.923 & 0.929 & 0.740 & 0.676 & 0.700 \\
      bert-large-uncased     & 0.929 & 0.936 & 0.933 & 0.846 & 0.790 & 0.817 \\
      deberta-v3-base    & \textbf{0.950} & 0.951 & 0.952 & 0.781 & 0.712 & 0.745 \\
      deberta-v3-large   & 0.936 & 0.932 & 0.939 & 0.854 & 0.766 & 0.813 \\
      MathBERT               & 0.928 & 0.930 & 0.938 & 0.811 & 0.761 & 0.765 \\
      \bottomrule
    \end{tabular}%
  }
\end{table}

For domain classification, as Table \ref {tab:bert_finetune_comparison} indicates, \texttt{DeBERTa-v3-base} leads with an F1 score of 0.950, closely followed by \texttt{RoBERTa-large} (0.943) and \texttt{DeBERTa-v3-large} (0.936), underscoring their powerful self-attention and deeper architectures.  \texttt{MathBERT} and \texttt{BERT-large} also achieve good results with F1 scores of 0.928 and 0.929 respectively.  In contrast, \texttt{ALBERT-base} (0.885) and \texttt{ELECTRA-small} (0.869) underperform, likely due to reduced parameter expressiveness and smaller model capacity, respectively. Both precisions and recalls are larger than 0.5 for all models.  

For skill classification, as Table \ref {tab:bert_finetune_comparison} shows, performance drops overall for the more challenging 19-category skill alignment.  \texttt{RoBERTa-large} performs the best (0.869), with \texttt{DeBERTa-v3-base} (0.853) and \texttt{BERT-large} (0.846) closely behind.  \texttt{RoBERTa-base} (0.833) and \texttt{MathBERT} (0.811) also show strong transfer.  \texttt{ALBERT-base} (0.712) and \texttt{ELECTRA-small} (0.278) remain weakest, suggesting that limited-capacity models struggle more with fine-grained skill distinctions. Both precisions and recalls are larger than 0.5 for all models except for \texttt{ELECTRA-small}.

\subsection{Ensemble Learning}
We first evaluated a majority‐voting ensemble that aggregates the skill predictions of all fine‐tuned BERT‐family models.  The ensemble produced weighted‐average F1 scores of 0.8394 for skill alignment and 0.9281 for domain alignment.  While majority voting can reduce individual model variance by leveraging prediction diversity, its performance remains below that of the best-performing single model (that is, \texttt{RoBERTa-large} with an F1 score of 0.869 on skill alignment and \texttt{DeBERTa-v3-large} with an F1 score of 0.950 on domain alignment), suggesting that the majority vote‐based ensemble does not fully exploit the most discriminative features learned by the best Transformer models.  

We also constructed a stacking ensemble model for skill classification using two base learners, an MLP trained on E5 embeddings (F1 = 0.847) and \texttt{RoBERTa-large} fine‐tuned for skill alignment (F1 = 0.869), and compared five meta‐models.  Table~\ref{tab:stacking_results} reports the weighted‐average F1 score for each meta‐model.

\begin{table}[ht]
  \caption{Stacking Ensemble Model Performance with Different Meta Models.}
  \label{tab:stacking_results}
  \centering
  \begin{tabular}{lr}
    \hline
    \textbf{Meta Model}          & \textbf{Weighted‐avg F1 Score} \\
    \hline
    Logistic regression          & 0.2013 \\
    Random forest                & 0.8405\\
    Gradient boosting            & 0.8433\\ 
    XGBoost                      & 0.8374 \\
    LightGBM                     & \textbf{0.8444} \\    
    \hline
  \end{tabular}
\end{table}

Among the meta‐models, \texttt{LightGBM} achieves the highest F1 score of 0.8444, marginally outperforming gradient boosting (0.8433) and random forest (0.8405).  
The worst performance by logistic regression (0.2013) indicates that a purely linear aggregator cannot capture the nonlinear dependencies between the base‐model predictions.  \texttt{LightGBM}'s superior performance can be attributed to its gradient‐based tree construction and regularization techniques, which effectively model complex interactions in the base‐prediction feature space. Despite the LightGBM meta‐model achieving a weighted‐average F1 of 0.844, the stacking ensemble  still underperforms the standalone \texttt{RoBERTa-large} model (F1 = 0.869).  

\section{Discussion}
\subsection{Embedding-based Models}
Our exploration of embedding‐based approaches for math item alignment reveals critical insights into the strengths and limitations of cosine similarity, supervised classification on raw embeddings, and linear dimensionality reduction.  
First, the mean‐pooling of the 1024‐dimensional E5 Sentence-Transformer embeddings within each of the nineteen skill categories and computing the full 19×19 cosine similarity matrix yielded off‐diagonal values uniformly above 0.88 (mean approximates 0.95). Such high similarity scores reflect the inherently homogeneous lexicon of mathematics problems: numerals, operators, common function names (e.g.\ “log”, “sin”), and standard phrases (e.g.\ “in terms of”, “perimeter of”) recur across skill labels.  

This lexical and syntactic overlap induces dense clustering in embedding space, undermining the capacity of cosine similarity to distinguish fine-grained content standards differences \citep{ButterfussDoran2025,ZhouOstrow2022}.  
Moreover, instruction‐tuned embedding models like E5 are optimized for semantic retrieval and zero‐shot transfer, not for delineating narrowly defined pedagogical categories without supervision. Consequently, cosine similarity metrics generate ambiguous nearest‐neighbor lists with extensive label overlap, precluding reliable automatic item alignment in the absence of downstream training.

Second, recasting the task as supervised classification on raw embeddings demonstrates the adaptability of high‐dimensional representations. We trained both linear classifiers (logistic regression, support vector machines) and nonlinear learners (multilayer perceptrons, gradient‐boosted trees) on the concatenated item attributes—stem, choices, rationale, and textual descriptions of tables and figures. Neural network-based MLPs achieved a weighted‐average F1 score of approximately 0.847 on the nineteen‐way skill classification, outperforming SVMs (0.800) and logistic regression (0.608).  

This performance hierarchy underscores that MLPs effectively carve nonlinear decision boundaries in the embedding manifold, integrating complex interactions among dimensions to isolate subtle semantic cues.  
By contrast, linear models struggle with noise and multicollinearity inherent in raw E5 embeddings, leading to suboptimal margin construction. Error‐analysis further revealed that linear classifiers often confuse adjacent skill labels with overlapping vocabulary (e.g.\ “Systems of two linear equations” vs. “Linear inequalities”), whereas MLPs more accurately capture contextual modifiers and positional patterns within the concatenated text.

Third, inspired by Gui (2025), we investigated PCA as dimension reduction to enhance the performance of linear learners. 
By projecting the 1024‐dimensional embeddings onto orthogonal axes of maximal variance, PCA denoises the feature space and mitigates multicollinearity \citep{Jolliffe2002,Abdi2010}.  
Under the Kaiser rule (eigenvalues > 1), logistic regression’s weighted‐average F1 scores increased from 0.608 to 0.811, and SVM’s F1 scores increased from 0.800 to 0.835. Similar improvements were observed at cumulative variance thresholds of 90 \% and 95 \%, indicating that a relatively small number of principal components capture most of the discriminative information for linear separation.  

However, PCA’s linear orthogonality inherently discards higher‐order interactions among embedding dimensions, as the principal axes are constrained to capture only the largest variance directions. This limitation became evident when applying the same reduced representations to MLPs: their F1 scores declined by over 0.03, illustrating that deep neural networks account for the richer, nonlinear interplay of dimensions that PCA filtering attenuates.

Collectively, these findings elucidate a fundamental trade‐off in embedding‐based item alignment: cosine similarity is too coarse for fine‐grained skill alignment; supervised learning on raw embeddings succeeds only with sufficiently expressive models; and linear dimensionality reduction benefits simpler learners at the expense of deep architectures.  

To bridge this gap, future work could explore nonlinear dimensionality reduction techniques such as variational autoencoders \citep{Hinton2006}, manifold learning methods like t-SNE \citep{Maaten2008} or UMAP \citep{McInnes2018} that aim to denoise embeddings while preserving intrinsic nonlinearity and hierarchical structures.  
Additionally, incorporating multi‐modal encoders capable of processing structured tables, figures, and LaTeX‐encoded equations may introduce orthogonal cues that alleviate lexical overlap. Finally, contrastive fine‐tuning objectives that explicitly push apart embeddings from different skill categories could augment separation in embedding space prior to PCA or classification \citep{Chen2020}.  
By balancing denoising with the retention of rich feature manifolds, such approaches promise more robust, accurate automated alignment of assessment items to educational content standards.

\subsection{BERT and Its Variant Model Fine Tuning}
Prior work on fine‐tuning transformers for  educational item alignment has largely concentrated on optimizing individual model performance for specific BERT‐family variants, typically evaluating a single architecture such as \texttt{bert-base-uncased} or RoBERTa on alignment tasks \citep{TanKim2024,Ding2025}. These studies have demonstrated that fine‐tuning pretrained transformers yields substantial gains over static embedding baselines, with F1 score improvements of 10–20 percentage points in many cases.  

Despite these successes, comparative evaluations of multiple BERT‐family models under a unified fine‐tuning protocol remain scarce.  Our study addresses this gap by systematically fine‐tuning eleven transformer models including \texttt{bert-base-uncased}, \texttt{bert-large-uncased}, \texttt{roberta-base}, \texttt{roberta-large}, \texttt{albert-base-v2}, \texttt{microsoft/deberta-v3-base}, \texttt{microsoft/deberta-v3-large}, \texttt{tbs17/MathBERT}, \texttt{google/electra-base-discriminator}, \texttt{google/electra-small-discriminator}, and \texttt{conv-bert-base} on the same math dataset with the same hyperparameter settings (learning rate, batch size, sequence length, early stopping criteria).  
This rigorous side-by-side comparison reveals consistent trends across both domain (four‐way) and skill (nineteen‐way) classification tasks.

Our results clearly underscore the superiority of larger, more sophisticated transformer models.  
\texttt{DeBERTa-v3-base} attains the highest weighted‐average F1 scores of 0.950 for domain alignment, closely followed by \texttt{roberta-large} (0.943) and \texttt{deberta-v3-large} (0.936). On the other hand, \texttt{roberta-large} attains the highest weighted‐average F1 scores of 0.869 for skill alignment, closely followed by \texttt{deberta-v3-large} (0.854) and \texttt{BERT-large-uncased} (0.846).  
These models benefit from deep architectures (12–24 transformer layers), extensive self‐attention heads (12–16 heads per layer), and enhanced pretraining objectives.  
In particular, DeBERTa’s disentangled attention mechanism, where content and position information are encoded separately, enables more precise modeling of the complex, formula‐rich text typical of math items, improving representation of spatial and relational cues in problems \citep{He2021}.  
Similarly, RoBERTa’s robust pretraining regimen, which omits the next‐sentence prediction task and employs dynamic masking over a larger corpus, enhances its contextual understanding of mathematical terminology and problem structures \citep{Liu2019}.

Task‐specific pretraining also confers advantages: \texttt{MathBERT}, pretrained on a large corpus of mathematical text (including LaTeX‐encoded formulas and problem annotations), achieves competitive performance, demonstrating the value of domain‐adapted language modeling for capturing the nuances of educational content. In contrast, lighter models such as \texttt{albert-base-v2}, which uses parameter‐sharing to reduce model size, and \texttt{google/electra-small-discriminator}, which employs a smaller generator/discriminator architecture, underperform. Their reduced parameter budgets and simplified objectives limit their capacity to model long‐range dependencies and the hierarchical complexity of math problems, resulting in lower discriminative power for fine‐grained labeling.

In addition to alignment accuracy comparisons, error‐analysis reveals that larger models more accurately classify items with rich context (e.g.\ word‐problems involving multiple steps, embedded figures or tables), whereas smaller models often fail on items requiring integration of spatial and algebraic reasoning.  
This suggests that attention depth and parameter capacity are critical for modeling multi‐clause prompts and implicit problem scaffolding. Moreover, we observe that model confidence distributions differ markedly. Larger models produce sharper probability peaks for correct labels, facilitating more reliable ensemble integration and potential calibration strategies, while smaller models yield flatter distributions that complicate threshold‐based decision rules.

From a practical standpoint, these findings inform model selection for high‐stakes assessment programs. While large transformers (e.g.\ DeBERTa‐v3‐base) deliver superior accuracy, their inference latency and GPU memory requirements are substantially higher, which may be prohibitive in resource‐constrained settings. Mid‐sized variants (e.g.\ \texttt{roberta-base}, \texttt{bert-large-uncased}) offer favorable trade‐offs, achieving near state‐of‐the‐art performance with reduced computational overhead.  
Finally, specialized models like MathBERT highlight the importance of domain‐adapted pretraining, suggesting that future efforts could invest in creating content‐aligned corpora to further enhance transformer efficacy on educational alignment tasks.

This comprehensive comparison demonstrates that, for automated content alignment in math items, larger transformer models with advanced attention mechanisms and domain‐adapted pretraining yield the best model performance. These results provide actionable guidance for assessment developers seeking to deploy robust, scalable alignment solutions, while also identifying directions for future research in efficient fine‐tuning, model compression, and content‐specific pretraining to balance accuracy with practical constraints.

\subsection{Ensemble Learning}
In this study, we investigated two ensemble strategies: majority voting and stacking, to leverage the strengths of diverse classifiers for math item alignment.  
Our majority‐voting ensemble aggregated the skill predictions of all fine‐tuned BERT‐family variants, yielding weighted‐average F1 scores of 0.8394 on the 19‐way skill alignment and 0.9281 on the 4‐way domain alignment. Although these scores did not exceed those of the top single model (\texttt{RoBERTa-large and DeBERTa-v3-base}, with F1 scores of 0.869 and 0.950, respectively on skill and domain alignment), majority voting demonstrated its utility in reducing prediction variance and stabilizing outputs across classes. This approach is straightforward to implement and requires no additional training, making it particularly attractive in production settings where inference speed and simplicity are paramount \citep{Dietterich2000}. In contexts where individual model failures are uncorrelated, majority voting can effectively correct outlier predictions by relying on consensus; however, in our experiments, the base models, sharing similar Transformer architectures and pretraining corpora, exhibited highly correlated errors, limiting the ensemble’s corrective capacity and resulting in only modest gains over weaker individual models.

To address the limitations of uniform vote weighting, we developed a stacking ensemble that combines two complementary base models: an MLP trained on instruction‐tuned E5 embeddings (F1 score of 0.847) and a fine‐tuned \texttt{RoBERTa-large} classifier (F1 score of 0.869). The multiple meta‐models including LightGBM, gradient boosting, random forests, logistic regression, and XGBoost was trained on the validation‐set probability outputs of both base models using k‐fold cross‐validation to avoid overfitting \citep{Wolpert1992}.  Among these meta‐models, LightGBM performed best, achieving a weighted‐average F1 score of 0.8444 for skill alignment and a F1 score of 0.9281 for domain alignment. The stacking’s ability to learn optimal weights and nonlinear combinations of the base predictions allowed it to correct systematic biases of individual models, such as overconfidence on particular skill subsets.  
Notably, the meta-model using logistic regression performed poorly (a F1 score of 0.201), underscoring the necessity of non‐linear meta‐models for capturing complex dependencies between base outputs.

Despite these methodological differences, both ensemble strategies share a core contribution: they enhance prediction robustness by diversifying error sources.  
Majority voting enhances the influence of any single model’s misclassifications, while stacking exploits complementary capacities between embedding‐based and Transformer‐based representations. Together, they provide a layered defense against both random and systematic errors, leading to more consistent alignment across heterogeneous content labels, including multi‐step word problems, table‐driven questions, and graph‐based prompts.  
This robustness is critical in high‐stakes assessment programs, where misalignment of items can propagate into flawed curriculum alignment and misinformed instructional decisions.

From the standpoint of operational deployment, the two approaches offer distinct trade‐offs. Majority voting’s minimal overhead makes it ideal for low‐latency pipelines, as it only requires parallel inference of base models followed by a simple argmax aggregation. In contrast, stacking demands additional computation for meta‐model training and inference, as well as careful management of validation splits to generate reliable meta‐features.  
However, stacking’s controlled training process enables practitioners to train ensemble models via hyperparameter tuning of the meta‐model, providing a balance between interpretability (via feature importances in tree‐based models) and predictive power.  
In resource‐constrained environments, a hybrid approach applying majority voting among a subset of lightweight models and reserving stacking for critical or ambiguous items may yield practical efficiency without sacrificing accuracy.

Looking forward, our findings highlight several avenues for enhancing ensemble-based labeling systems. First, introducing weighted voting schemes, where each model’s vote is scaled by its validation‐set performance or class‐specific precision, could improve majority‐voting efficacy by amplifying reliable predictors \citep{Sill2009}. Second, expanding the base‐model pool to include orthogonal architectures such as static‐embedding models, sequence‐to‐sequence encoders, or multimodal networks specialized for tables and figures would increase representation diversity and strengthen the ensemble’s corrective capacities \citep{Dietterich2000}. Third, enriching the feature space for the meta-model in stacking by incorporating per‐class confidence scores, intermediate layer activations, token‐length statistics, and PCA‐based summary features could provide the meta‐model with deeper insights into prediction uncertainty and item complexity. Finally, model distillation techniques could be applied post‐ensemble to compress the behavior of the full ensemble into a single, efficient student model, retaining ensemble-level accuracy while minimizing inference cost \citep{Hinton2015}.

Our ensemble learning experiments demonstrate that both majority voting and stacking are valuable tools for improving the accuracy of item content alignment in math. By carefully selecting ensemble strategies that align with operational constraints and by incorporating future refinements such as weighted aggregation, model diversity, and richer meta‐model features, it is expected that more robust and scalable systems for item alignment can be built.

\section{Conclusions}
This study evaluated three approaches to align Math items to both domains and skills: (1) embedding‐based classification using instruction‐tuned E5 embeddings, which provided moderate baseline performance and, when combined with PCA, substantially boosted linear classifiers; (2) end‐to‐end fine‐tuning of diverse BERT‐family models, where \texttt{DeBERTa-v3-base} achieved the highest weighted‐average F1 scores on domain alignment and \texttt{RoBERTa‐large} achieved the highest weighted‐average F1 scores on skill alignment; and (3) ensemble learning, where majority voting yielded good but not better results and stacking with LightGBM marginally improved skill classification but did not surpass the best single BERT related model. These findings underscore the superior effectiveness of large, instruction‐tuned Transformers for fine‐grained item alignment, while highlighting the potential of PCA for enhancing linear model performance and the importance of base‐model diversity and feature richness for training meta‐models  in ensemble learning.  

This study is subject to several limitations. First, the dataset comprises only 1,385 math items, which may restrict the generalizability of our findings to other subject domains or assessment formats.  Second, non‐textual information (tables, figures, equations) was converted into plain text descriptions, potentially omitting visual or spatial cues that could benefit specialized models.  Third, we truncated inputs to 512 tokens for Transformer fine‐tuning, which may have discarded important long‐range context in the longest items.  Fourth, our PCA analysis employed only linear dimensionality reduction, leaving unexplored nonlinear techniques (e.g.\ autoencoders) that might better preserve complex feature interactions.  Finally, the stacking ensemble was limited by the small number and high correlation of base models; broader model diversity and richer features for meta-model training may be necessary to realize the full potential of ensemble methods. Future work could explore nonlinear embedding compression (e.g.\ autoencoders), specialized encoders for tables and figures, and hybrid ensembling strategies with performance‐weighted voting or neural meta‐models to further advance automated item alignment.  

Though the rapid evolution of large language models (LLMs) such as OpenAI’s ChatGPT and Google’s Gemini can be a feasible approach to item alignment, its adoption for item alignment in large-scale high-stakes assessment programs remains challenging given test security concerns. Accordingly, all model training in this study was conducted on local, on-premise infrastructure rather than via cloud-based platforms.  Though this localized approach ensures strict data privacy and aligns with test security protocols of test stakeholders, it may limit access to the latest advances in the LLM architectures and the scalable computing resources offered by cloud services. Future studies could explore the trade-off between test security and technology innovation brought by AI for automated item alignment in large-scale high-stakes assessment programs.

\printbibliography
\end{document}